\documentclass{article}

\usepackage[preprint]{nips_2018}

\usepackage{graphicx}
\usepackage{subcaption}
\usepackage[export]{adjustbox}
\usepackage{algorithm}
\usepackage[noend]{algpseudocode}
\usepackage{setspace}
\usepackage{amsmath}

\usepackage[utf8]{inputenc} 
\usepackage[T1]{fontenc}    
\usepackage{hyperref}       
\usepackage{url}            
\usepackage{booktabs}       
\usepackage{amsfonts}       
\usepackage{nicefrac}       
\usepackage{microtype}      

\title{An End-to-end Approach to Semantic Segmentation with 3D CNN and Posterior-CRF in Medical Images}

\author{
  Shuai Chen \\
  Erasmus MC\\
  \texttt{s.chen.2@erasmusmc.nl} \\
  \And
  Marleen de Bruijne \\
  Erasmus MC\\
  \texttt{marleen.debruijne@erasmusmc.nl } \\
}

\begin{document}

\maketitle

\begin{abstract}
Fully-connected Conditional Random Field (CRF) is often used as post-processing to refine voxel classification results by encouraging spatial coherence. In this paper, we propose a new end-to-end training method called Posterior-CRF. In contrast with previous approaches which use the original image intensity in the CRF, our approach applies 3D, fully connected CRF to the posterior probabilities from a CNN and optimizes both CNN and CRF together. The experiments on white matter hyperintensities segmentation demonstrate that our method outperforms CNN, post-processing CRF and different end-to-end training CRF approaches.

\end{abstract}

\section{Introduction}

Conditional random fields have been widely used as an efficient post-processing method in medical image segmentation [1][2][11]. There is a more elegant way to take advantage of CNN and CRF at the same time, which has been introduced the first time by Zheng et al.[3] to train 2D CNN and CRF together, and further developed by Monteiro et al.[6] into a 3D version. There are also other similar joint optimization of CNN and CRF approaches investigated by different researchers [4][5].

However, the current end-to-end training methods in medical imaging still rely on independent tuning of some of the CRF parameters and all use intensity information as the primary feature space. In medical images, intensity information often provides low-quality feature space for the CRF as intensities are noisy and several structures belonging to different classes may have the same intensity. To counter this, we propose a new CRF method called Posterior-CRF that can be applied at the end of a segmentation CNN (such as U-net). In contrast with previous end-to end approaches, it optimizes all CRF parameters during network training and applies the CRF to the posterior probability map instead of the original intensity information. In this way, the mean field inference in CRF could make full use of the high-quality feature maps obtained with CNN. The experiments show that our approach outperforms the post-processing CRF and previous end-to-end CRF approaches in white matter hyperintensities segmentation.

\section{Methods}
\label{methods}

\subsection{CNN modeling}

\begin{figure}[h]
\centering
\includegraphics[height=3.8cm]{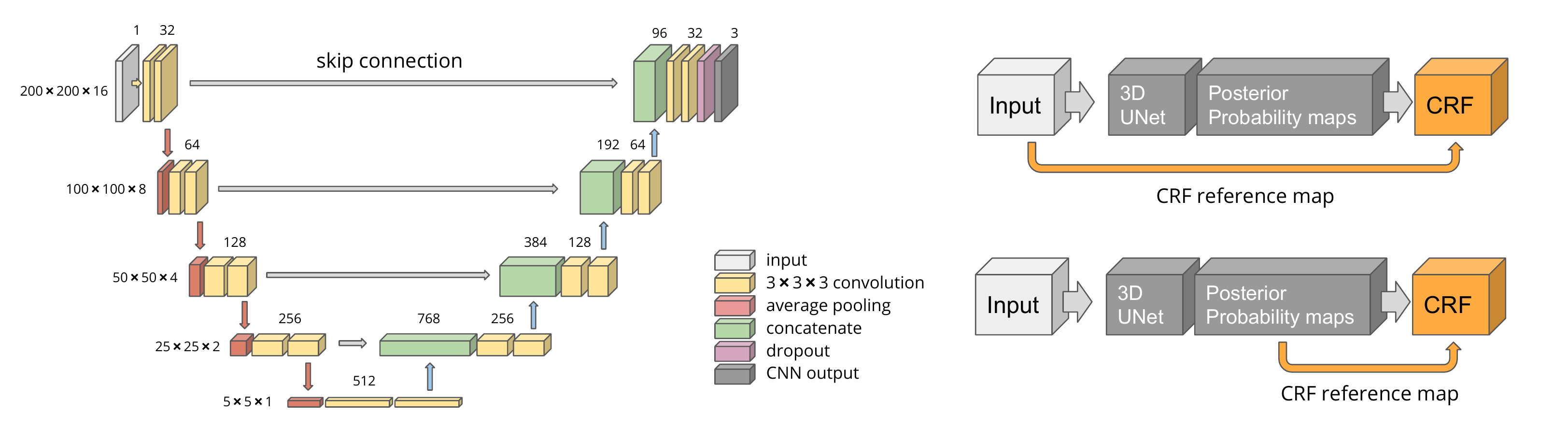}
\caption{\textbf{End-to-end training networks.} For each graph: 3D UNet baseline (left), Intensity-CRF (upper right) and Posterior-CRF Neural Network (lower right).}
\label{fig:network}
\end{figure}

We use 3D UNet [7] as the baseline architecture in this paper. The details of the network can be found in Fig \ref{fig:network}. 
All convolution layers in UNet use \textit{ReLU} as activation function except for the last output layer, which use \textit{softmax} to produce the final CNN probability maps. We use categorical cross-entropy as the loss function.

\subsection{Posterior-CRF}

In the fully-connected CRF model ($\mathbf{X}$,$\mathbf{I}$), the corresponding Gibbs energy w.r.t the label segmentation $\mathbf{x}$ is

\begin{equation}
\label{Gibbsenergy}
E({\mathbf{X}=\mathbf{x}|\mathbf{I}}) = \sum_{i} \varphi_u(x_i|\mathbf{I}) + \sum_{i<j} \varphi_p(x_i,x_j|\mathbf{I})
\end{equation}

where $i$ and $j$ range from 1 to $N$, which is the number of voxels in the random field $\mathbf{X}$ and 3D input patch $\mathbf{I}$. For convenience, the conditioning on $\mathbf{I}$ will be omitted in the rest of the paper. The first term $\varphi_u(x_i)$ is the unary potential, which in our case is the current voxelwise class probabilities in the last CNN layer. The second term $\varphi_p(x_i,x_j)$ is the pairwise potential:

\begin{equation}
\label{pairwise}
\varphi_p(x_i,x_j) = \mu(x_i,x_j)[ \omega^{(1)}\mathbf{exp}(-\frac{|p_i-p_j|^2}{2\theta_{\alpha}^2}-\frac{|f_i-f_j|^2}{2\theta_{\beta}^2}) 
+ \omega^{(2)}\mathbf{exp}(-\frac{|p_i-p_j|^2}{2\theta_{\gamma}^2})]
\end{equation}

where $\mu(x_i,x_j)$ is the label compatibility function given by Potts model $\mu(x_i,x_j)=[x_i\neq{x_j}]$ that captures the compatibility between different pairs of labels. $\omega$ is the linear combination weight of different predefined kernels.

The first kernel in Eq \ref{pairwise} is defined by the positions vectors $p_i$ and $p_j$ and feature vectors $f_i$ and $f_j$, and the second kernel is the \textit{smoothness kernel} which is only controlled by the voxel positions. $\theta_\alpha$, $\theta_\beta$ and $\theta_\gamma$ are the parameters that control the sensitivity to the corresponding feature space. In the previous methods, people usually use the intensity $I$ of the input image as the feature (or reference map) $f$, which we call Intensity-CRF methods (Fig \ref{fig:network}). However, the Intensity-CRF is very sensitive to the parameter $\theta_{\beta}$ because the intensity varies a lot between different medical images as well as the random noise. Therefore, we replace the intensity $I$ by the posterior probability $x$ as the new reference maps, which we call Posterior-CRF method (Fig \ref{fig:network}). The idea of Posterior-CRF is to try to use the best-quality CNN feature maps as the feature space used in the mean field inference. As an efficient feature extractor, 3D UNet could provide high-quality feature maps which provide better class separation compared to the original input image. Moreover, Posterior-CRF also avoids the noisy intensity feature space that makes the inference in Intensity-CRF unstable. Another advantage of Posterior-CRF is that compared with Intensity-based methods, there is no longer the need to pretrain and fix the CRF parameter settings because we don't use original intensity information anymore. And now, all the parameters like $\omega^{(1)}$, $\omega^{(2)}$, $\theta_{\alpha}$, $\theta_{\beta}$, $\theta_{\gamma}$ are equivalently trained together with the other weights in the network.

\section{Experiments}

We test our methods on 60 FLAIR scans from \textit{WMH 2017 Challenge} [9]. The images were acquired from three hospitals and manually annotated with three labels: background, white matter hyperintensities and other pathology. Images are randomly split into 36 images for training, 12 for validation and 12 for testing. Training patches are extracted and cropped to(or padded if it is smaller than) the size $ 200 \times 200 \times 16$ with the original voxel size ($ 0.96 \times 0.96 \times 3.00$, $ 1.00 \times 1.00 \times 3.00$, $ 1.20 \times 0.98 \times 3.00$ $m^{3}$ for three hospitals) and intensities, with 87.5$\%$ overlap in z-direction between patches. 
Several 3D data augmentation strategies were applied on the training patches, including 3D rotation with randomly sampled from [$10^{\circ}$, $5^{\circ}$, $5^{\circ}$], shifting by [7, 24, 24] voxels, as well as flipping in all 3 directions (XY, XZ and YZ). 
We trained our network on all the three labels and report the results of white matter hyperintensities versus background + other pathology.

\begin{table}[h]
  \caption{\textbf{Results of WMH 2017 dataset} (DSC: Dice similarity coefficient. H95: Hausdorff distance (95th percentile). AVD: Average volume difference. FP: number of False positive voxels. FN: number of False negative voxels).}
  \label{wmhresults}
  \centering
  \begin{tabular}{llllll}
  
    \toprule
   
    Method     & DSC(std)    & H95(std)(mm) & AVD(std)($\%$) & FP(std) & FN(std)\\
    \midrule
    UNet     & 0.683(0.068) & \textbf{4.637}(1.374)  & 42.58(10.07) & 5952(1238.07) & \textbf{514}(337.16)\\
    Post-CRF     & 0.676(0.096) & 7.955(1.371)  & 75.37(8.99) & 5622(1558.38) & 754(148.29)\\
    Intensity-CRF   & 0.682(0.087)   &   4.773(1.542)  & 41.56(13.19) & 5730(985.98) & 569(511.45) \\
    Spatial-CRF     & 0.707(0.081)    &  8.741(3.144) & 49.83(19.71) & \textbf{701}(364.67) & 3490(1300.44)\\
    Posterior-CRF & \textbf{0.747}(0.064) &  5.513(2.818)  & \textbf{21.80}(5.92) & 3387(507.41) & 1118(559.66)\\
    
    \bottomrule
  \end{tabular}
\end{table}

\begin{figure}[h]
\centering
\includegraphics[height=6.0cm]{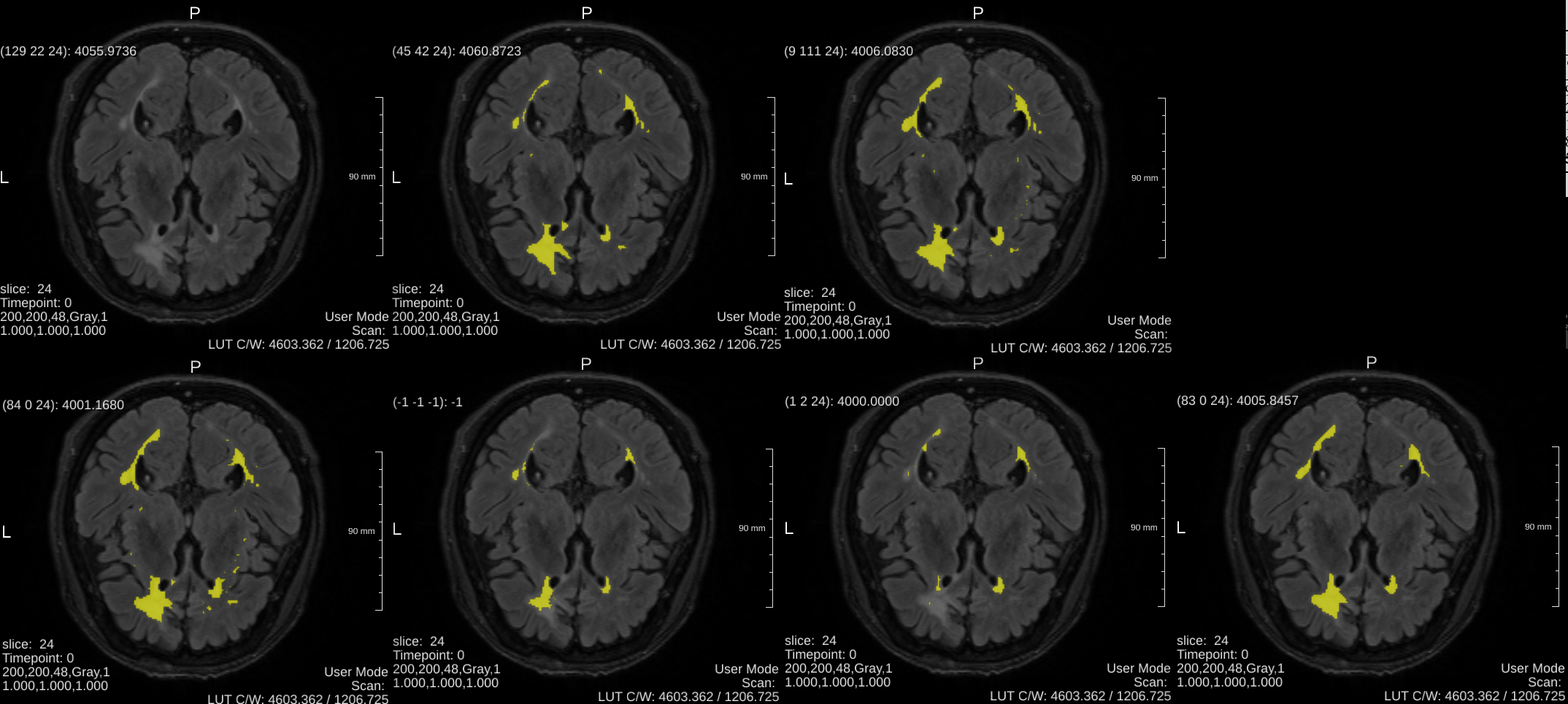}
\caption{\textbf{Visual comparison of different methods.} \textit{Upper row}: For each figure: (1) Original FLAIR image (2) Manual annotation (3) 3D UNet baseline. \textit{Lower row}: (4) Post-CRF (5)Intensity-CRF (6) Spatial-CRF and (7) Posterior-CRF. Better to view with zooming in color.}
\label{fig:WMH}
\end{figure}

As shown in Table \ref{wmhresults}, we compared our method with UNet and three different CRF approaches, which are: Post-CRF[3] for post-processing, Intensity-CRF[4] and Spatial-CRF that only use the position information (second term in Eq \ref{pairwise}). Parameters $\omega$ and $\theta$ for Post-CRF are tuned by grid search on the training sets. For Intensity-CRF and Spatial-CRF, the relevant $\theta$ are taken from Post-CRF, while the weights $\omega$ are learned. Posterior-CRF achieves the best Dice score and Average volume difference while it also performs well on Hausdorff distance and has a good balance between FP and FN. For the visualization of a certain slice results in Fig \ref{fig:WMH}, we can see that Posterior-CRF has the best visual quality compared to other approaches. The UNet and the Post-CRF results are visually equivalent and have many false positives, while Intensity-CRF and Spatial-CRF are similar and both remove too many voxels.

\section{Conclusion}

We propose a new end-to-end CRF approach called Posterior-CRF that could be trained together with CNN in a better way and overcome the drawbacks of other CRF approaches. 

\section*{References}

\small

[1] Kamnitsas, K., Ledig, C., Newcombe, V. F., Simpson, J. P., Kane, A. D., Menon, D. K., ... $\&$ Glocker, B. (2017). Efficient multi-scale 3D CNN with fully connected CRF for accurate brain lesion segmentation. Medical image analysis, 36, 61-78.

[2] Dou, Q., Yu, L., Chen, H., Jin, Y., Yang, X., Qin, J., $\&$ Heng, P. A. (2017). 3D deeply supervised network for automated segmentation of volumetric medical images. Medical image analysis, 41, 40-54.

[3] Zheng, S., Jayasumana, S., Romera-Paredes, B., Vineet, V., Su, Z., Du, D., ... $\&$ Torr, P. H. (2015). Conditional random fields as recurrent neural networks. In Proceedings of the IEEE international conference on computer vision (pp. 1529-1537).

[4] Schwing, A. G., $\&$ Urtasun, R. (2015). Fully connected deep structured networks. arXiv preprint arXiv:1503.02351.

[5] Lin, G., Shen, C., Van Den Hengel, A., $\&$ Reid, I. (2016). Efficient piecewise training of deep structured models for semantic segmentation. In Proceedings of the IEEE Conference on Computer Vision and Pattern Recognition (pp. 3194-3203).

[6] Monteiro, M., Figueiredo, M. A., $\&$ Oliveira, A. L. (2018). Conditional Random Fields as Recurrent Neural Networks for 3D Medical Imaging Segmentation. arXiv preprint arXiv:1807.07464.

[7] Ronneberger, O., Fischer, P., $\&$ Brox, T. (2015). U-net: Convolutional networks for biomedical image segmentation. In International Conference on Medical image computing and computer-assisted intervention (pp. 234-241). Springer, Cham.

[8] Krähenbühl, P., $\&$ Koltun, V. (2011). Efficient inference in fully connected crfs with gaussian edge potentials. In Advances in neural information processing systems (pp. 109-117).

[9] http://wmh.isi.uu.nl/

[10] Adams, A., Baek, J., $\&$ Davis, M. A. (2010). Fast high dimensional filtering using the permutohedral lattice. In Computer Graphics Forum (Vol. 29, No. 2, pp. 753-762). Oxford, UK: Blackwell Publishing Ltd.

[11] Wachinger, C., Reuter, M., $\&$ Klein, T. (2018). DeepNAT: Deep convolutional neural network for segmenting neuroanatomy. NeuroImage, 170, 434-445.

\end{document}